\documentclass[letterpaper]{article}
\usepackage[preprint]{aaai2027}
\nocopyright  
\usepackage[hyphens]{url}
\usepackage{graphicx}
\usepackage{natbib}
\usepackage{caption}
\usepackage{booktabs}
\usepackage{amsmath}
\usepackage{amssymb}
\usepackage{array}

\title{NormWorlds-CF: Solver-Verified Counterfactual Normative Reasoning\\with Metamorphic-Relation GRPO}
\author{
    Xinqi Zhang\textsuperscript{\rm 1}
}
\affiliations{
    \textsuperscript{\rm 1}Tsinghua University
}

\newcommand{\method}{MR-GRPO}
\newcommand{\dataset}{NormWorlds-CF}
\newcommand{\lab}[1]{\textsc{#1}}

\begin{document}

\maketitle

\begin{abstract}
Language models can reach the right normative verdict for the wrong reason. We introduce \dataset, a solver-verified environment for counterfactual normative reasoning in executable rule worlds. Its deterministic solver produces final answers, proof and falsification certificates, argument statuses, support sets, and paired-world change labels, enabling supervision and evaluation without LLM judges. The benchmark contains staged SFT diagnostics and a compact paired-world task with 270 root families and 1080 canonical-to-variant pairs. The SFT diagnostics show that final-answer supervision can saturate verdict accuracy without inducing falsification competence: answer-only SFT reaches perfect answer accuracy but scores zero on joint falsification certificates, while full-mix training with targeted replay reaches strong all-task accuracy (0.99). For the structured-change task, we introduce metamorphic-relation GRPO (\method), a class-conditioned reward for GRPO that gives partial credit for relation families and solver-visible change fields. In matched Qwen3-1.7B continuation experiments, \method{} improves held-out relation accuracy and relation-family correctness, and reduces wrong-family error, compared to sparse and answer-only GRPO. In Qwen3-4B three-seed validation, sparse reward preserves coarse relation labels best, answer-only reward improves answer-change but weakens relation-family structure, and \method{} leads on answer-, support-, and status-change fields as well as class-conditioned MR and change-presence. These results show that verified counterfactual structure can shape post-training beyond final answers, while exact full change-record generation, invariant subtype recognition, and out-of-distribution (OOD) transfer remain open problems.
\end{abstract}

\section{Introduction}

Normative reasoning exposes a failure mode that ordinary post-training metrics miss: a model can be right at the surface and wrong in the structure. A useful reasoner in legal, institutional, or policy-like settings must track rules, exceptions, priorities, defeated arguments, and counterfactual edits \cite{vonwright1951deontic,nute1994defeasible,prakken2015law,guha2023legalbench}. Predicting \lab{permitted} is useful only if the model can name the decisive rule, explain why a plausible prohibition is defeated, and recognize when removing a priority edge should flip the conclusion. Final-answer accuracy---and therefore final-answer reward---can hide exactly these failures.

This problem is easy to understate if one imports the dominant post-training template from mathematics. In math word problems and contest-style reasoning, supervision and reward often reduce to a sparse final answer: the boxed number or option is a strong proxy for correctness, and process rewards are valuable mainly because the answer space is vast and lucky guessing is rare \cite{cobbe2021gsm8k,lightman2023letsverify,shao2024deepseekmath}. Normative decision making is different in kind. The ``answer'' is typically drawn from a small label set (\lab{permitted}/\lab{forbidden}/\lab{obligated}/\ldots), so high answer accuracy can be achieved by shallow cues. What matters is dialectical: which rule survives, which conflicting argument is defeated, and whether a proposed justification should be rejected. In that sense, our setting is closer to a structured judgment task than to answer-sparse mathematical derivation---and answer-only supervision is insufficient and potentially misleading as a proxy for dialectical competence.

The empirical bottleneck is therefore attribution, not a shortage of benchmarks. Legal evaluation suites and legal-domain LLMs are essential for external validity \cite{chalkidis2022lexglue,guha2023legalbench,fei2024lawbench,colombo2024saullm}, but realistic legal tasks entangle knowledge, retrieval, drafting, and reasoning. Controlled reasoning benchmarks isolate compositionality \cite{weston2015babi,johnson2017clevr,clark2020transformers,tafjord2021proofwriter}, yet rarely expose defeasible conflict, priority, and defeated-argument structure. Post-training research needs environments that are rich enough to be interesting and executable enough that supervision, reward, and evaluation do not depend on LLM judges---whose own normative judgments would reintroduce the very attribution problem we aim to study.

\begin{figure*}[t]
\centering
\includegraphics[width=0.92\textwidth]{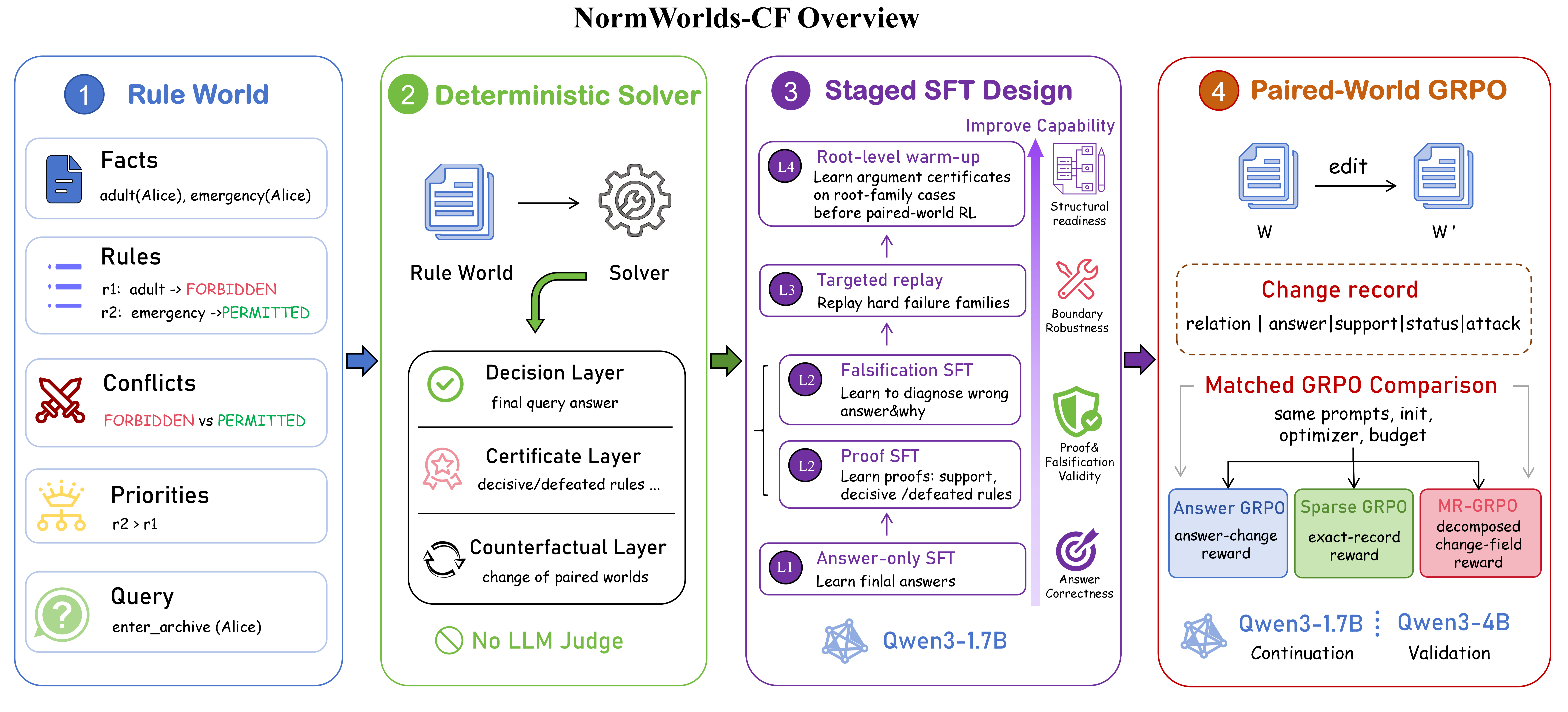}
\caption{Overview of \dataset. Executable rule worlds are solved by deterministic procedures that produce verified answers, certificates, and paired-world change records. These solver-generated structures support staged SFT diagnostics and a matched GRPO comparison among answer-only, sparse exact/schema, and MR-aware change-field rewards---all evaluated by deterministic scorers, without LLM judges.}
\label{fig:overview}
\end{figure*}

We address this gap with \dataset, a solver-verified laboratory for counterfactual normative reasoning (Figure~\ref{fig:overview}). Each instance is generated from an executable rule world; deterministic solvers emit answers, proof and falsification certificates, argument statuses, minimal support, and root-level metamorphic relations. Errors become inspectable: answer error, invalid proof, missing falsification, relation-family confusion, schema failure, or counterfactual generalization failure can be separated. The construction is motivated by a concrete design bet: if we want to study \emph{how} post-training shapes normative structure, we must first make that structure machine-checkable.

The study is staged by design. We first ask whether small models can learn solver-generated interfaces under SFT, comparing answer-only, answer-plus-proof (no falsification), answer-plus-proof-plus-falsification, and targeted-replay curricula. Only after structured outputs stabilize do we move to RL; otherwise an RL failure would be uninterpretable. For RL, we formulate a compact paired-world change task inspired by metamorphic testing \cite{chen1998metamorphic,chen2018metamorphic}: each root family pairs a canonical world with transformed variants that should preserve or change normative structure in named ways. The model outputs a machine-scored change record over relation, answer, support, status, and attack fields---harder than final-answer prediction, but more focused than regenerating two full certificates.

We compare answer-only reward, sparse exact/schema GRPO \cite{shao2024deepseekmath,deepseekai2025r1}, and \method{}, a class-conditioned reward that gives partial credit for metamorphic structure. The central empirical finding is a \emph{reward geometry}: different rewards induce different shortcuts rather than a single winner. Answer-only reward repairs local answer-change fields while shifting the relation prior toward change labels. Sparse reward preserves coarse relation labels but often leaves many training groups with zero within-group reward variance, starving GRPO's ranking signal. \method{} densifies that signal and improves answer-, support-, and status-change fields together with class-conditioned soft scores, while exact full-record generation, invariant subtype recognition, and OOD transfer remain open.

\subsubsection{Contributions}
This paper makes three contributions:
\begin{enumerate}
\item We introduce \dataset, a solver-verified environment for counterfactual normative reasoning with executable worlds, deterministic certificates, compact paired-world change records, family-level splits, and OOD structural probes---motivated by the need to separate normative structure from open-text legal confounding.
\item We show that final-answer supervision can saturate verdict accuracy without inducing solver-verified falsification competence: proof and falsification expose different capabilities, and targeted replay repairs concrete failures under deterministic scoring.
\item We formulate compact structured-change prediction as a reward-ready post-training task and introduce \method. Under matched protocols, answer-only, sparse, and MR-aware GRPO learn different slices of the same structure and expose characteristic shortcuts.
\end{enumerate}

The organizing claim is broader than ``MR-GRPO beats a baseline.'' Outcome supervision cannot identify normative structure quality; solver-verified counterfactual structure makes that quality measurable at three levels---supervision targets, evaluation tasks, and reward design---especially when answers are low-entropy and structure is scarce. Scope is controlled: \dataset{} is a laboratory for verified normative structure, not a substitute for open-ended legal expertise.

\section{NormWorlds-CF}

\dataset{} separates three objects often conflated in natural-language benchmarks: a symbolic world (source of truth), a textual rendering (prompt only), and solver-generated labels (never an LLM judge).

\subsection{Design Principles: Why Build It This Way}

The dataset is not a miniaturized legal corpus. It is built to answer a post-training question that open-text legal benchmarks cannot cleanly ask: when a model improves, \emph{what} improved? Three principles follow.

\subsubsection{Executable worlds before fluent text}
Realistic statutes mix doctrine, drafting conventions, and retrieval. We instead start from ground facts, typed rules, and priorities whose consequences are computable. Templates may generate surface wording, but ground truth never depends on that wording. This reverses a common pipeline that annotates natural text first and hopes structure can be recovered later.

\subsubsection{No LLM judges by construction}
If labels or rewards were themselves produced by an LLM, evaluation would inherit that model's normative biases and its tendency to reward fluent but structurally wrong justifications. Here every answer, certificate field, falsification error type, and metamorphic change record is emitted by a deterministic solver over a closed DSL. Scoring is likewise parser-based. The advantage is not only reproducibility: it makes reward design scientifically meaningful, because credit assignment refers to the same objects the evaluator checks.

\subsubsection{Dialectical targets, not answer-sparse math}
Math RL often treats a rare correct final answer as the learning signal. Our answers are dense and low-entropy; the scarce resource is \emph{structural fidelity}. Call this a \emph{scarcity inversion}: in contest math, the bottleneck is finding the answer; in normative judgment, the bottleneck is defending and attacking under conflict. That is why the task suite foregrounds proof (justify the winner), falsification (reject a plausible loser), and paired-world change (state what a transformation should alter). These are judgment-style interfaces: the model must accept or reject candidate structure under an executable oracle, rather than search a vast answer space for a single token sequence.

\subsection{Executable Worlds and Certificates}

A world is a finite rule program $w=(F,R,P,q)$ with ground facts $F$, instantiated typed rules $R$, priority relation $P\subseteq R\times R$, and query $q$. Templates may be used during generation, but the solver receives only ground rule objects: each rule has an identifier, a kind, an action, and a finite set of preconditions. A priority edge $(r_i,r_j)\in P$ means that $r_i$ overrides $r_j$ under conflict. A deterministic solver $S$ maps $w$ to an answer $a$ and certificate $C$:
\begin{equation}
S(w)=(a,C),
\end{equation}
where $C$ contains decisive rules, defeated alternatives, argument statuses, and minimal support. If no applicable normative rule supports the query, the answer is \lab{no\_conclusion}; if incompatible undefeated arguments remain, \lab{conflict\_unresolved}; otherwise the answer is read from surviving labels under a fixed reporting convention (\lab{obligated} before \lab{forbidden} before \lab{permitted}). That convention is only a serialization order, not a claim about legal or ethical priority. Transformations $\tau$ are executable edits. Each $\tau$ carries a family label $r_{\tau}$ determined by its edit operator and validated against solver-computed pre- and postconditions; fine-grained answer, support, status, and attack changes are obtained by differencing the two solver traces,
\begin{equation}
M(w,\tau)=\big(r_{\tau},\,D\big(S(w),S(\tau(w))\big)\big),
\end{equation}
not by inspecting free-form text after the fact. In particular, $r_{\tau}$ distinguishes relation subtypes that share similar or empty field diffs (e.g., \lab{surface\_invariant} vs.\ \lab{distractor\_invariant}).

Consider a toy archive-entry case: facts \lab{apprentice(alice)} and \lab{emergency\_rescue(alice)}; rule r1 forbids entry for apprentices; rule r2 permits entry under emergency rescue; priority r2 > r1. Both rules apply and conflict; priority selects \lab{permitted}, with decisive rule r2 and defeated rule r1. Removing the priority edge yields \lab{conflict\_unresolved}; removing the emergency fact yields \lab{forbidden}; adding an irrelevant fact should leave the certificate invariant. These edits define the four root-level metamorphic families used later. The same object also makes reward differences concrete: an output that flips the answer-change field but invents a change relation on an invariant edit can be credited by answer-only reward, penalized by sparse exactness, and only partly credited by \method{} through family and emptiness terms.

Label construction is a closed symbolic procedure over a fixed schema: forward-chain identity rules to a fixed point; select applicable normative arguments whose preconditions hold; attack only incompatible labels (permission vs.\ prohibition, obligation vs.\ prohibition); apply priorities by directed graph reachability; then extract certificates and falsification error types (e.g., defeated rule cited as decisive, unresolved conflict ignored, missing precondition). Minimal support searches subsets of decisive rules sufficient to sustain the answer while defeating conflicting alternatives. Supervision therefore has three layers: a \emph{decision} layer (final answer), a \emph{certificate} layer (proof, falsification, statuses, support), and a \emph{counterfactual} layer (paired-world relation and change fields). In the archive example, predicting \lab{forbidden} is an answer error; predicting \lab{permitted} while citing r1 is a proof error; failing to reject candidate \lab{forbidden} is a falsification error; claiming that priority removal changes nothing is a counterfactual error. This typed failure taxonomy is itself a contribution of the environment: it turns ``the model is wrong'' into a researchable diagnosis.

\subsection{Diagnostic Task Ladder}

The same worlds feed a staged interface suite (Table~\ref{tab:tasks}) rather than a single monolithic benchmark. The result is a diagnostic ladder: a model may learn final answers before proofs, proofs before falsification, and still fail when asked to compare a root world with a transformed variant. Answer and certificate tasks establish whether solver structure is learnable under SFT. Early micro corpora make those interfaces measurable (sizes and task mixes in the supplementary material). Surface-varied micro data (1000 roots, 4000 instances) then varies lexical forms, entity names, rule-id prefixes, and fact/rule order while preserving solver-equivalent structure, checking that models learn the rule pattern rather than a single template; all renderings of one root stay in the same split.

The compact change-record task is the RL bridge. Regenerating two nested certificates proved too noisy in pilot prompts---a failed output could reflect reasoning failure, formatting failure, or excessive sequence length---so the model emits a five-field record (Table~\ref{tab:schema}): relation plus answer/support/status/attack changes. This target is small enough to score repeatedly during GRPO, yet structured enough to distinguish coarse relation recognition from fine-grained change description.

\begin{table}[t]
\centering
\setlength{\tabcolsep}{3pt}
\small
\begin{tabular}{@{}p{0.20\linewidth}p{0.72\linewidth}@{}}
\toprule
Interface & What it tests \\
\midrule
Answer & Final decision on original or edited worlds \\
Proof & Decisive/defeated rules and support facts \\
Falsification & Reject invalid answers/certificates with error type \\
Arg.\ cert. & Accepted/defeated/undecided argument nodes \\
Change rec. & Relation + answer/support/status/attack changes \\
\bottomrule
\end{tabular}
\caption{Task interfaces derived from the same solver-verified worlds.}
\label{tab:tasks}
\end{table}

\begin{table}[t]
\centering
\setlength{\tabcolsep}{2pt}
\small
\begin{tabular}{@{}>{\raggedright\arraybackslash}p{0.24\linewidth}>{\raggedright\arraybackslash}p{0.68\linewidth}@{}}
\toprule
Field & Scored content \\
\midrule
relation & One of four MR labels \\
answer change & Unchanged/changed; both answers \\
support change & Added/removed rules, facts, priorities \\
status change & Added/removed argument statuses \\
attack change & Added/removed attack edges \\
\bottomrule
\end{tabular}
\caption{Compact paired-world change-record schema.}
\label{tab:schema}
\end{table}

Root-level metamorphic families are the main counterfactual substrate (Table~\ref{tab:scale}). Each family anchors one canonical world and four variants with distinct diagnostic roles: \lab{surface\_invariant} (reorder facts/rules/priorities only; identifiers unchanged), \lab{support\_delta} (edit a decisive support fact), \lab{priority\_delta} (edit priority edges), and \lab{distractor\_invariant} (add irrelevant facts/rules for an unrelated action). The canonical world itself is an unscored anchor; supervised examples are the four canonical-to-variant pairs. Family types target residual failures seen in earlier stages---partial priority conflicts, unresolved conflicts, multiple compatible overrides, missing-precondition distractors---rather than arbitrary noise. Splits are by root family, so variants never leak across train/dev/ID/OOD. OOD uses held-out structural family types (e.g., priority chains and distractor-conflict variants), not merely new roots of the same type mix. Critically, the relation label names the \emph{mechanism} of change, not a fixed answer direction: in some families both support and priority edits resolve an unresolved conflict into \lab{permitted}, while in others they remove a conclusion or introduce a new one. That is the point of the construction---to force models to explain \emph{why} two worlds differ, not only whether their final labels differ.

\begin{table}[t]
\centering
\small
\begin{tabular}{lrrrl}
\toprule
Split & Fam. & Cases & Pairs & Role \\
\midrule
Train & 120 & 600 & 480 & Optimization \\
Dev & 30 & 150 & 120 & Diagnostics \\
ID test & 60 & 300 & 240 & Held-out ID \\
OOD test & 60 & 300 & 240 & Held-out types \\
\bottomrule
\end{tabular}
\caption{Root-level MR pilot scale (270 families, 1080 pairs).}
\label{tab:scale}
\end{table}

\section{Post-Training Method}

We use LoRA adapters throughout \cite{hu2021lora}. SFT establishes structured-output competence; RL then compares rewards on the compact change-record task under matched data, optimizer, and group size. The experimental variable is the reward geometry. We report two complementary protocols: a Qwen3-1.7B continuation that starts from a shared sparse-GRPO adapter, and a Qwen3-4B scale check that starts from a shared aligned-SFT adapter.

\subsection{SFT as Interface Substrate}

SFT is part of the method, not a disposable warmup. Because the task is judgment-structured rather than answer-sparse, we treat interface competence as a scientific prerequisite: if a model cannot emit proof or falsification fields under direct supervision, later RL comparisons would confound reward design with serialization failure.

Answer-only training uses only original- and edited-world answer targets. The answer-plus-proof condition keeps those answer targets and adds decisive rules, defeated rules, and support---but no falsification. Answer-plus-proof-plus-falsification keeps the answer and proof targets and further adds falsification certificates (error type and counter-evidence). Targeted replay keeps that full mix while oversampling residual failure families (unresolved-vs-priority conflicts, identity-derived support, sibling counter-rule confusions). A short root-level case warmup transfers argument-certificate competence to the MR family distribution before RL.

\subsection{GRPO and Three Rewards}

For prompt $x$, GRPO samples a group of $G$ completions and normalizes rewards within the group \cite{shao2024deepseekmath}:
\begin{equation}
A_i=\frac{r_i-\frac{1}{G}\sum_{j}r_j}{\mathrm{std}(\{r_j\})+\epsilon}.
\end{equation}
We compare three rewards that instantiate three post-training philosophies. Answer-only reward is the natural import of math-style outcome supervision:
\begin{equation}
R_{\mathrm{ans}}=\omega_{\mathrm{ans}}A_{\mathrm{ans}}+\omega_{\mathrm{schema}}S_{\mathrm{ans}}+\omega_{\mathrm{top}}S_{\mathrm{top}}+R_{\mathrm{len}}.
\end{equation}
It scores answer-change correctness plus minimal format shaping, and ignores relation, support, status, and attack structure. Sparse reward instead demands exact compact-record generation:
\begin{equation}
R_{\mathrm{sparse}}=\lambda_{\mathrm{all}}A_{\mathrm{all}}+\lambda_{\mathrm{full}}S_{\mathrm{full}}+\lambda_{\mathrm{top}}S_{\mathrm{top}}+\lambda_{\mathrm{opt}}O_r+R_{\mathrm{len}},
\end{equation}
where $A_{\mathrm{all}}$ is strict all-fields correctness. It is a strong exactness baseline, but provides little signal when a completion is structurally close yet inexact. \method{} replaces both with a class-conditioned metamorphic reward. Relations are grouped into invariant relations $\mathcal{I}=\{\lab{surface\_invariant},\lab{distractor\_invariant}\}$ and change relations $\mathcal{C}=\{\lab{support\_delta},\lab{priority\_delta}\}$:
\begin{align}
R_{\mathrm{MR}} &= R_{\mathrm{rel}}+R_{\mathrm{schema}}+g_{\Delta}R_{\mathrm{sem}}+R_{\mathrm{len}}, \nonumber\\
g_{\Delta} &= \gamma_0+\gamma_{\Delta}S_{\Delta}, \nonumber\\
R_{\mathrm{rel}} &= \alpha_r A_r+\alpha_{\mathrm{fam}}A_{\mathrm{fam}} \nonumber\\
&\quad +\alpha_{\mathrm{opt}}O_r-\alpha_{\mathrm{wrong}}W_{\mathrm{fam}}.
\end{align}
Here $A_r$ is exact relation correctness, $A_{\mathrm{fam}}$ is invariant-vs-change family correctness, and $W_{\mathrm{fam}}$ penalizes crossing that boundary. Semantic terms give exact and partial credit on answer/support/status/attack fields and on a softer change-presence score that checks nonempty change slots without requiring exact item identities. The class-conditioned term is the design centerpiece:
\begin{equation}
A_{\mathrm{class}}=
\begin{cases}
\eta^{I}_{\mathrm{fam}}A_{\mathrm{fam}}+\eta_{\mathrm{same}}A_{\mathrm{same}}+\sum \eta E, & r^\star\in\mathcal{I},\\
\eta^{C}_{\mathrm{fam}}A_{\mathrm{fam}}+\eta_{\mathrm{chg}}A_{\mathrm{chg}}+\eta_{\mathrm{prs}}A_{\mathrm{prs}}, & r^\star\in\mathcal{C},
\end{cases}
\end{equation}
so invariant gold labels reward unchanged answers and empty change fields, while change gold labels reward changed-answer structure and appropriate change presence. Soft scores reuse these solver-visible fields; exact relation and relation-family accuracy remain the hard structural checks that a dense reward need not win. Coefficients are fixed before held-out evaluation (full weights in the supplementary material); controls are protocol-matched rather than coefficient-count matched, because answer-only must remain an outcome-only control. The intended experimental difference is therefore reward geometry, not optimizer, data, initialization, or budget.

\begin{table}[t]
\centering
\small
\begin{tabular}{llll}
\toprule
Setting & Init & Reward & Steps \\
\midrule
Sparse & sparse60 & exact/schema & +60 \\
Answer-only & sparse60 & answer-change & +60 \\
\method{} & sparse60 & class-cond.\ MR & +60 \\
\bottomrule
\end{tabular}
\caption{Qwen3-1.7B matched continuation from a shared sparse60 adapter. Qwen3-4B uses a shared aligned-SFT start (three seeds, 40 steps).}
\label{tab:match}
\end{table}

Table~\ref{tab:match} summarizes the Qwen3-1.7B continuation: all three branches continue from the same sparse-GRPO adapter for $+60$ steps on relation-balanced change records, so the comparison isolates reward switching after a shared sparse phase. The Qwen3-4B validation \cite{yang2025qwen3} asks the same three-way reward question from a shared aligned-SFT initializer---three seeds, a matched 40-step budget, and only the reward differing within each seed---providing a cleaner from-SFT check at larger scale (seeds and hardware in the supplementary material).

\section{Experiments}

The experiments answer four questions. Does final-answer supervision suffice for solver-generated normative structure? After a shared sparse phase, how do answer-only, sparse, and MR-aware continuations diverge on Qwen3-1.7B? From a shared aligned-SFT start on Qwen3-4B, does the same reward geometry reappear when change fields are more expressible? Which failures remain?

All metrics use deterministic parsers and scorers---never an LLM judge. We treat \emph{relation}, \emph{relation-family}, \emph{wrong-family}, and exact answer/support/status change as primary structural readouts. Soft scores (\emph{change-presence}, \emph{class-conditioned MR}) reuse solver-visible fields and can align with the \method{} reward; they are secondary diagnostics, not substitutes for exact family checks. Formally: \emph{relation} is four-way exact accuracy; \emph{relation-family} is invariant-vs-change correctness; \emph{wrong-family} crosses that boundary; \emph{all-changes exact} requires the four change fields to match, excluding the relation label; \emph{change-presence} checks answer kind and nonempty slots without exact item identities; \emph{class-conditioned MR} is $A_{\mathrm{class}}$. Predicting \lab{priority\_delta} for gold \lab{support\_delta} is wrong on subtype but right on family; predicting an invariant label for that case is a wrong-family error. Unless noted, the SFT surface ablations and the Qwen3-1.7B GRPO continuation are single runs; the Qwen3-4B GRPO study averages three independent seeds (seeds, hardware, and per-seed tables in the supplementary material). Exact 0/1 metrics use McNemar tests on the 1.7B held-out pairs; soft scores are secondary means. OOD asks whether in-distribution gains leave the training relation mix.

\subsection{SFT Diagnostics}

Table~\ref{tab:sft} shows the surface-variation ablation. Mixes are independent from-base runs (896 records, 280 steps); the last row keeps the full mix with targeted failure-family oversampling. Answer-only SFT saturates Answer (original- and edited-world single-world answers) but scores zero on joint falsification. The prompts already name the falsification task and schema, and models emit partially correct fields (Answer wrong-rule 0.56; Answer+Proof counter-rule 0.42) while the joint certificate stays empty---so the result shows limited zero-shot transfer from answer/proof supervision to exact falsification certificates under the current interface, not only a formatting failure (field breakdown in the supplementary material). Answer+Proof saturates proofs yet still scores zero on joint falsification. Adding falsification yields the first strong all-task jump; full mix + targeted replay reaches 0.99 overall. On root-level MR cases, a strong initializer still scores only 0.075 all-fields correct before a 600-step argument-certificate warmup raises this to 0.775, leaving relation-sensitive errors for RL.

\begin{table}[t]
\centering
\setlength{\tabcolsep}{5pt}
\small
\begin{tabular}{@{}lrrrr@{}}
\toprule
Targets & All$\uparrow$ & Answer$\uparrow$ & Proof$\uparrow$ & Fals$\uparrow$ \\
\midrule
None (prompt only) & 0.02 & 0.04 & 0.00 & 0.00 \\
Answer & 0.53 & 1.00 & 0.10 & 0.00 \\
Answer+Proof & 0.75 & 0.99 & 1.00 & 0.00 \\
Answer+Proof+Fals. & 0.97 & 0.98 & 0.97 & 0.95 \\
Full mix + targeted replay & 0.99 & 1.00 & 0.98 & 0.99 \\
\bottomrule
\end{tabular}
\caption{Surface-variation SFT diagnostics (896 / 280 from base). Rows 2--4 ablate the target mix; the last adds targeted failure-family oversampling. Answer = mean of original- and edited-world single-world answer accuracy (same schema, different input worlds); Proof/Fals: certificate exactness; All: mean of those two answer tasks plus proof and falsification.}
\label{tab:sft}
\end{table}

\subsection{Matched Qwen3-1.7B Continuation}

On the 240-example held-out ID test (60 per relation), continuing with \method{} improves relation accuracy from 0.2417 to 0.3375, relation-family correctness from 0.2917 to 0.3750, and wrong-family error from 0.5750 to 0.4333 over continuing with sparse reward (Table~\ref{tab:1p7b}; McNemar $p=6.06\times10^{-4}$, $4.53\times10^{-3}$, $8.22\times10^{-6}$ on held-out items). Gains concentrate on change-family labels: \lab{priority\_delta} rises from 39/60 to 48/60 and \lab{support\_delta} from 18/60 to 33/60, while both invariant subtypes remain near floor. Exact full change-record metrics also remain near floor; at this scale we read the continuation as evidence about relation-family learning after a shared sparse phase, not as exact field generation. Answer-only continuation is not stronger than sparse here: it ties on answer-change (0.05) and is worse on relation-family behavior. Dev checks (dev32 and stratified40) match the held-out direction.

\begin{table}[t]
\centering
\small
\begin{tabular}{lrrr}
\toprule
Metric & Sparse & Ans-only & \method{} \\
\midrule
Relation$\uparrow$ & 0.242 & 0.221 & \textbf{0.338} \\
Answer change$\uparrow$ & 0.050 & 0.050 & 0.038 \\
Relation-family$\uparrow$ & 0.292 & 0.267 & \textbf{0.375} \\
Wrong-family$\downarrow$ & 0.575 & 0.625 & \textbf{0.433} \\
Class-cond.\ MR$\uparrow$ & 0.302 & 0.326 & \textbf{0.335} \\
\bottomrule
\end{tabular}
\caption{Held-out ID Qwen3-1.7B three-way reward comparison.}
\label{tab:1p7b}
\end{table}

\subsection{Qwen3-4B Scale Validation}

From a shared aligned-SFT start, the same geometry reappears more cleanly (Table~\ref{tab:4b}; three-seed means on the shared 240-example held-out set). On primary exact metrics, sparse GRPO preserves coarse relation and relation-family best (0.608 / 0.790). Answer-only GRPO improves answer-change from 0.610 to 0.743 over sparse, and also lifts secondary change-presence---but relation-family drops from 0.790 to 0.703 and wrong-family rises from 0.210 to 0.297. Error slices show the shortcut: answer-only reward shifts the relation prior toward change labels (e.g., predicting \lab{support\_delta} far more often), helping changed-answer cases while converting invariant examples into the change family---the risk of importing outcome-only supervision into a judgment task. \method{} leads on exact answer-, support-, and status-change fields and on the secondary soft scores, while staying close to sparse on relation-family and far above answer-only; relative to answer-only, it improves mean relation and relation-family in every seed.

\begin{table}[t]
\centering
\small
\begin{tabular}{lrrr}
\toprule
Metric & Sparse & Ans-only & \method{} \\
\midrule
Relation$\uparrow$ & \textbf{0.608} & 0.540 & 0.594 \\
Answer change$\uparrow$ & 0.610 & 0.743 & \textbf{0.747} \\
Support change$\uparrow$ & 0.562 & 0.597 & \textbf{0.621} \\
Status change$\uparrow$ & 0.571 & 0.565 & \textbf{0.600} \\
All changes exact$\uparrow$ & 0.431 & 0.435 & \textbf{0.443} \\
Relation-family$\uparrow$ & \textbf{0.790} & 0.703 & 0.781 \\
Wrong-family$\downarrow$ & \textbf{0.210} & 0.297 & 0.219 \\
Class-cond.\ MR$\uparrow$ & 0.768 & 0.759 & \textbf{0.800} \\
Change-presence$\uparrow$ & 0.813 & 0.829 & \textbf{0.843} \\
\bottomrule
\end{tabular}
\caption{Qwen3-4B three-seed mean on held-out test240.}
\label{tab:4b}
\end{table}

Training diagnostics explain the mechanism (Table~\ref{tab:train-diag}). Sparse exact reward leaves about 44\% of logged GRPO groups with zero within-group reward standard deviation, starving the ranking signal---a failure mode familiar from sparse outcome rewards, but here induced by exact nested structure rather than by a rare boxed answer. Answer-only densifies that signal by rewarding answer-change alone. \method{} drops zero-std groups to about 11\% while keeping the highest training-batch class-conditioned MR probe, matching the held-out pattern of denser structural credit without collapsing to the most visible answer field.

\begin{table}[t]
\centering
\small
\begin{tabular}{lrr}
\toprule
Branch & Zero-std$\downarrow$ & Class-MR probe$\uparrow$ \\
\midrule
Sparse & 0.442 & 0.674 \\
Answer-only & 0.317 & 0.693 \\
\method{} & \textbf{0.108} & \textbf{0.698} \\
\bottomrule
\end{tabular}
\caption{Qwen3-4B training diagnostics (three-seed mean). Zero-std: fraction of logged groups with zero reward std.; Class-MR probe: mean $A_{\mathrm{class}}$ on training completions.}
\label{tab:train-diag}
\end{table}

\subsubsection{How to read the three-way comparison}
The claim is a geometry, not a ranking. Answer-only chases the most visible local field and shifts the relation prior toward change labels. Sparse preserves coarse relation labels but often yields no within-group ranking signal for near-miss structure. \method{} densifies credit along the invariant/change boundary and solver-visible change fields. Sparse winning exact relation/family and answer-only winning answer-local fields are expected signatures of those regimes, not contradictions.

\subsection{Boundaries and Negative Results}

OOD remains open: on Qwen3-1.7B OOD240, relation accuracy ties at 0.258; \method{} reduces wrong-family error (0.667$\rightarrow$0.571) without improving overall OOD relation or class-conditioned MR. Exact full-record generation and invariant subtype recognition (\lab{surface\_invariant} vs.\ \lab{distractor\_invariant}) stay weak even when models detect that nothing substantive changed. Invariant-heavy reweighting and family-pooled advantage normalization do not close these gaps. The positive claim is therefore in-distribution structural attribution under matched rewards, not transfer or full-record mastery.

\section{Discussion: Insights and Implications}

\subsubsection{Scarcity inversion relative to math post-training}
In contest-style math, a rare correct final answer is often informative \cite{cobbe2021gsm8k,lightman2023letsverify,shao2024deepseekmath}. In normative judgment the label set is tiny, so scarcity moves from \emph{which answer} to \emph{which structure}. That is why answer-only SFT can look solved and why answer-only GRPO can ``improve'' by biasing toward change labels while dialectical checks remain empty.

\subsubsection{Proof supervision does not transfer to falsification}
Proof exhibits the winner; falsification rejects a plausible loser. Our ladder shows limited automatic transfer under deterministic scoring: answer-plus-proof can saturate proofs while joint falsification stays at zero. Scoring only the winning path misses certificate-review errors---citing a defeated rule, ignoring unresolved conflict, or accepting an invalid certificate.

\subsubsection{Interventions, partial credit, and attribution}
Root-level MR pairs are interventions, not generic augmentations: surface/distractor edits should leave certificates fixed, while support/priority edits should change them in solver-specified ways. Compact change records are a practical RL interface for naming the mechanism of change. Confusing \lab{support\_delta} with \lab{priority\_delta} is wrong on subtype yet right on the invariant/change boundary; class-conditioned partial credit follows that diagnostic hierarchy, whereas sparse exactness ignores it and often starves GRPO. Refusing LLM judges keeps reward design an experimental variable, and typed failures make competence-level attribution possible. \dataset{} complements LegalBench-style evaluation by asking whether verified structure can be taught under auditable oracles before asking how much transfers to open legal text.

\section{Related Work}

We situate \dataset{} in three nearby literatures below. For a short primer, \citet{prakken2015law} surveys defeasible legal argumentation, \citet{tafjord2021proofwriter} exemplifies proof-oriented synthetic reasoning, and \citet{shao2024deepseekmath} introduces the GRPO-style group updates used in our RL stage (LoRA adapters follow \cite{hu2021lora}).

\subsubsection{Synthetic and solver-verified reasoning}
Controlled benchmarks isolate compositionality and rule following \cite{weston2015babi,johnson2017clevr,clark2020transformers,tafjord2021proofwriter}. \citet{sanyal2022robustlr} further test whether deductive models stay consistent under minimal logical edits to operators and theories. \dataset{} is complementary: rather than entailment labels over fixed theories alone, it exposes proof, falsification, and paired-world change under executable priorities, with solver-checked certificates rather than LLM-judged text.

\subsubsection{Normative and legal evaluation}
Deontic and defeasible logics emphasize why a conclusion follows under rules, exceptions, and priorities \cite{vonwright1951deontic,nute1994defeasible,prakken2015law}. Recent LLM evaluations probe defeasible property inheritance at scale \cite{allaway2025defreasing}, normative versus epistemic modal reasoning patterns \cite{ozeki2025normative}, and diagnostic legal tasks with machine-computable statute encodings \cite{servantez2026openexempt}. Legal evaluation suites and legal-domain LLMs measure external validity on realistic texts and workflows \cite{chalkidis2022lexglue,guha2023legalbench,fei2024lawbench,colombo2024saullm}. We ask a complementary post-training question: can reward design teach verified normative structure under controlled transformations?

\subsubsection{Process supervision, metamorphic tests, and counterfactuals}
Intermediate and verifiable supervision improve beyond final answers in mathematics and preference learning \cite{cobbe2021gsm8k,lightman2023letsverify,christiano2017deep,ouyang2022training,shao2024deepseekmath,deepseekai2025r1}. Rubric-structured rewards provide multi-criterion partial credit when binary oracles are scarce, usually scored by an LLM judge \cite{gunjal2025rubrics}; our rewards instead score solver-visible change fields without a judge. Metamorphic testing and counterfactual/contrast sets expose local decision failures when oracles are hard \cite{chen1998metamorphic,chen2018metamorphic,kaushik2020counterfactual,gardner2020contrast}; unlike generic augmentation aimed only at invariance, our priority and support edits are supposed to change answers and certificates in precise ways, and those relations become RL targets. Persistent strict serialization failures also motivate future constrained decoding \cite{scholak2021picard}.

\section{Conclusion}

\dataset{} makes counterfactual normative structure auditable: answers, certificates, and paired-world change labels are solver-generated rather than LLM-judged. SFT diagnostics show that final-answer supervision---often a reasonable default in answer-sparse math---can saturate verdicts without inducing falsification competence. On compact change records, matched answer-only, sparse, and MR-aware GRPO induce a clear geometry under a 1.7B continuation protocol and a 4B from-SFT protocol: sparse leads on coarse relation labels, answer-only trades family structure for answer-local gains, and \method{} leads on answer/support/status fields and secondary soft scores. Exact full-record generation, invariant subtype recognition, and OOD transfer remain open. The broader lesson is that outcome supervision cannot identify normative structure quality, so post-training should use verified structure at three levels: supervision targets, counterfactual tasks, and reward design.

\clearpage
\bibliography{references}

\end{document}